\documentclass{article}

\usepackage{arxiv}

\usepackage[utf8]{inputenc}
\usepackage[T1]{fontenc}
\usepackage[english]{babel}
\usepackage{microtype}
\usepackage{amsmath, amssymb, amsfonts}
\usepackage{mathtools}
\usepackage{booktabs}
\usepackage{multirow}
\usepackage{graphicx}
\usepackage[caption=false]{subfig}
\usepackage{float}
\usepackage{rotating}
\usepackage{url}
\usepackage{xcolor}
\usepackage{nicefrac}
\usepackage[ruled,vlined,linesnumbered]{algorithm2e}
\usepackage{hyperref}
\usepackage{cleveref}

\hypersetup{
  colorlinks = true,
  linkcolor  = black,
  citecolor  = black,
  urlcolor   = blue
}

\renewcommand{\headeright}{Preprint}
\renewcommand{\undertitle}{Preprint}
\renewcommand{\shorttitle}{Active-DiNTS: Active Differentiable Network Topology Search}

\title{Active-DiNTS: Active Differentiable Network Topology Search}

\author{%
  \textbf{Gean Trindade Pereira}\thanks{Corresponding author.} \\
  Institute of Mathematics and Computer Sciences (ICMC)\\
  University of S\~ao Paulo\\
  S\~ao Carlos, 13560-970, S\~ao Paulo, Brazil \\
  \texttt{geantrinpereira@gmail.com} \\
  \And
  \textbf{Thierry Urruty} \\
  XLIM, Universit\'e de Poitiers\\
  Universit\'e de Limoges, CNRS\\
  Poitiers, 86000, Vienne, France \\
  \texttt{thierry.urruty@univ-poitiers.fr} \\
  \AND
  \textbf{Muriel Visani} \\
  Laboratoire Informatique, Image et Interaction (L3i)\\
  La Rochelle Universit\'e\\
  La Rochelle, 17042, Charente-Maritime, France \\
  \texttt{muriel.visani@univ-lr.fr} \\
  \And
  \textbf{Andr\'e C. P. L. F. de Carvalho} \\
  Institute of Mathematics and Computer Sciences (ICMC)\\
  University of S\~ao Paulo\\
  S\~ao Carlos, 13560-970, S\~ao Paulo, Brazil \\
  \texttt{andre@icmc.usp.br} \\
}

\date{October 2026}

\begin{document}
\maketitle

\begin{abstract}
Neural Architecture Search (NAS) has proved to be a strong alternative to manual network design, but applying it to 3D medical image segmentation is limited by two well-known costs, large annotation budgets and multi-GPU clusters. Thus, this paper introduces \textbf{Active-DiNTS} (Active Differentiable Network Topology Search), an approach that embeds pool-based Active Learning (AL) into a bi-level differentiable topology search to perform architecture discovery and label curation jointly. At each query round, unlabeled MRI volumes are ranked by one of three uncertainty signals (Entropy, Variance, or Standard Deviation), and only the top-ranked volumes are sent to an oracle for annotation. The new labels feed two interlocked stages. Network weights are updated in an outer loop, while the macro/micro topology of a U-Net-style backbone is refined in an inner loop. Three AL regimes (\emph{weights-only}, \emph{topology-only}, \emph{joint}) expose the speed--accuracy trade-off. Evaluations on the Medical Segmentation Decathlon (MSD) Task01 BrainTumour benchmark showed that Active-DiNTS surpasses DiNTS, C2FNAS, and nnU-Net in Dice, with gains of about 10 percentage points on Edema and 5 points on Non-Enhancing core, on a single GPU and using a fraction of the labeled volumes. The discovered architectures are denser and more FLOP-heavy than prior baselines, but remain competitive in trainable parameters and peak memory; the fastest search variant finishes in under 0.25 GPU-days, over 27$\times$ faster than the eight-GPU DiNTS search. Together, these results indicate that pairing differentiable NAS with active data acquisition is a practical recipe for accurate 3D segmentation under realistic constraints.
\end{abstract}


\section{Introduction}
\label{sec:intro}

Volumetric segmentation underpins much of modern radiology software. It is what allows clinicians to outline tumors, plan radiotherapy on a per-organ basis, and follow lesions through time at voxel resolution~\cite{li2023transforming}. The workhorse for these tasks has been, for almost a decade, the U-Net family of convolutional encoder--decoder models~\cite{ronneberger2015u,milletari2016v,zhou2019unet++}. Strong as these models are, building a variant that suits a new dataset still relies on substantial manual tuning and domain knowledge.

A line of work known as Neural Architecture Search (NAS) tries to remove that handcrafted step by treating the network topology itself as something to be optimized~\cite{white2023neural,real2019regularized,ren2026automated}. Among NAS strategies, differentiable methods have been particularly impactful. By recasting discrete operator choices as continuous parameters that can be trained with gradient descent, they have brought search budgets down from thousands of GPU-hours to single-digit GPU-days~\cite{liudarts}. In medical imaging, DiNTS~\cite{he2021dints} pushed this idea further by searching over multi-resolution, multi-path encoder--decoder topologies, producing segmentation networks that match or beat well-tuned 3D U-Nets; a recent systematic review of differentiable NAS for medical segmentation~\cite{benedykciuk2026dnas} confirms that this line of work remains an active research direction in the area.

Yet a structural problem persists. The two implicit prerequisites of NAS, a fully annotated training corpus and a sizable GPU pool, rarely line up with the reality of clinical research groups. Pixel-level expert labeling of 3D scans is laborious, slow, and expensive, and the typical hospital research lab does not have a multi-node cluster on hand. That mismatch raises a natural question: \emph{is it possible to search for a strong architecture while also keeping the number of labeled volumes the search consumes as small as possible?}

Active Learning (AL) is a good fit for that question. AL frameworks query annotations only for the samples that the model is least confident about, and they have repeatedly demonstrated annotation savings across many supervised settings~\cite{cohn1994improving,cohn1996active,dasgupta2011two,ren2021survey}. For voxel-level prediction tasks, uncertainty signals such as per-voxel entropy can be aggregated into volume-level scores that drive the selection process, provided that some care is taken with outliers and class imbalance~\cite{leite:2010,wang2024deepal}.

Progress in medical image segmentation over the last two years has unfolded along two complementary tracks. One track scales \emph{model and data}. Vision foundation models such as the Segment Anything Model (SAM)~\cite{kirillov2023segment} and its medical adaptations~\cite{ma2024segment,wu2025medical}, together with transformer-based 3D backbones~\cite{hatamizadeh2022unetr,hatamizadeh2022swin,tang2022selfsupervised}, lean on large-scale pretraining and substantial compute to obtain strong general-purpose representations. The other track scales \emph{efficiency and adaptability}. Dataset-driven systems such as nnU-Net~\cite{isensee2019nnunet,isensee2024nnunet} configure themselves automatically to a given dataset without external pretraining. Active-DiNTS lives squarely in the second track and pushes it further by jointly tackling two constraints, scarce annotations and single-GPU search, that the first track does not, and was not designed to, address.

Our claim in this paper is that AL and NAS are mutually reinforcing rather than competing. AL feeds the search with high-value supervision; the search, in turn, shapes the model that will produce the next round of uncertainty estimates. Building on that observation, we propose \textbf{Active-DiNTS}, a differentiable topology search augmented with uncertainty-driven pool-based AL, and we show empirically that the two components compound. On MSD Task01 BrainTumour, the resulting pipeline reaches state-of-the-art Dice with a single GPU and a small slice of the annotation budget.

In summary, this study presents the following contributions:
\begin{itemize}
  \item A \textbf{one-shot differentiable topology search interleaved with pool-based AL} for 3D medical segmentation. Two coupled optimization loops, one for weights and one for topology, draw labeled samples from independent, uncertainty-ranked pools, which makes it possible to discover architectures with very few annotations.
  \item A \textbf{flexible AL engine} that supports three update regimes (\emph{weights-only}, \emph{topology-only}, \emph{joint}) and three uncertainty heuristics (Entropy, Variance, Standard Deviation), enabling explicit control over the speed--accuracy trade-off.
  \item \textbf{New best Dice scores} on MSD/Task01 BrainTumour relative to DiNTS, C2FNAS, nnU-Net, and other strong baselines, achieved on a single GPU and with substantially less labeling effort.
  \item A \textbf{resource-efficiency analysis} showing that the discovered networks, despite being denser and heavier in FLOPs, are competitive in trainable-parameter count and in peak GPU memory, and that the search itself is up to 27$\times$ faster than the DiNTS baseline.
\end{itemize}

\section{Background and Related Work}
\label{sec:related}

\subsection{U-Net Architectures for Medical Segmentation}

The U-Net design proposed by Ronneberger et al.~\cite{ronneberger2015u} has effectively defined what a baseline medical segmentation model looks like. Several extensions have followed: V-Net~\cite{milletari2016v} lifted the idea to volumetric inputs by adding residual links and a Dice-based objective, and UNet++~\cite{zhou2019unet++} reworked the skip connections to better fuse information across scales. For brain tumor segmentation, Multi-Planar U-Net (also distributed as CerebriuDIKU)~\cite{perslev2019one} sidesteps the memory cost of full 3D convolutions by combining a single 2D U-Net with multi-planar test-time augmentations. nnU-Net~\cite{isensee2019nnunet} takes yet another route, dispensing with manual tuning altogether. It automatically configures and ensembles a family of U-Net variants according to dataset-specific properties (resolution, voxel spacing, class frequencies). These hand-crafted or semi-automated solutions provide strong baselines but, by virtue of being topologically fixed, leave some accuracy on the table. A recent comprehensive review by Azad et al.~\cite{azad2024medical} surveys the broader U-Net ecosystem and shows how influential this family has remained.

On the volumetric side, \c{C}i\c{c}ek et al.~\cite{cciccek20163d} introduced a 3D U-Net that learns from sparsely annotated slices, and the Attention U-Net~\cite{oktay2018attention} showed that lightweight attention gates suffice to focus the decoder on regions of interest without requiring a separate detector. Among semi-supervised approaches, UMCT (sometimes labeled NVDLMED)~\cite{xia20203d} combines multi-view co-training with uncertainty-weighted label fusion, allowing the model to learn from both labeled and unlabeled examples.

\subsection{NAS for 3D Segmentation}

NAS applied to 3D medical segmentation is a comparatively young area. SCNAS~\cite{kim2019scalable} explores encoder--decoder structures via a stochastic continuous-relaxation scheme that scales to high-resolution input. C2FNAS~\cite{yu2020c2fnas} proceeds in two passes, first choosing a macro topology and then refining cell-level operators by one-shot uniform sampling, to reduce the discrepancy between the architecture found during search and the architecture actually deployed.

DiNTS~\cite{he2021dints} raised the bar by performing a fully differentiable search over multi-resolution, multi-path topologies. Through a feature-node conversion trick and a topology-aware loss term, DiNTS reduces the so-called \emph{discretization gap}, the inconsistency that arises when continuous architecture probabilities are rounded into a discrete graph. Even so, DiNTS shares two limitations with its NAS peers, both of which Active-DiNTS targets directly. The entire labeled training set is assumed to be available, and the search is run on a multi-GPU node (eight V100s in DiNTS's case). Despite the recent shift of part of the community toward transformer-based backbones and foundation-model fine-tuning (Sec.~\ref{sec:related-transformer}), differentiable NAS remains a practical option when neither large pretraining corpora nor multi-GPU resources are available, as the contemporary review in~\cite{benedykciuk2026dnas} also documents.

\subsection{Active Learning for Dense Prediction}

Pool-based AL keeps a large unlabeled reservoir and at each round picks the most informative samples to be annotated, according to an acquisition function~\cite{cohn1994improving,cohn1996active}. Standard heuristics include uncertainty sampling, query-by-committee, and expected model change~\cite{leite:2010}; recent surveys give a thorough account of how these classical ideas extend to the deep-learning setting~\cite{ren2021survey}, and Wang et al.~\cite{wang2024deepal} provide a dedicated treatment of deep AL in medical image analysis, including the specific difficulties that dense prediction raises for acquisition functions designed with classification in mind. For segmentation, uncertainty is typically computed per voxel, often as entropy of the predicted class distribution, and then condensed into a single score per volume that drives the ranking.

The integration of AL with model training has been studied extensively, but its combination with architecture search has received little attention. Active-DiNTS targets that gap: AL is wired directly into the bi-level optimization loop of a differentiable NAS so that the architecture and the labeled pool evolve together toward higher segmentation quality.

\subsection{Transformers and Foundation Models for Medical Segmentation}
\label{sec:related-transformer}

A second wave of progress in medical segmentation, parallel to NAS, has emerged from large-scale pretraining and attention-based architectures. UNETR~\cite{hatamizadeh2022unetr} and Swin UNETR~\cite{hatamizadeh2022swin} replace U-Net's convolutional path with self-attention blocks that capture long-range dependencies, and they benefit substantially from self-supervised pretraining on large unlabeled medical corpora~\cite{tang2022selfsupervised}. On the foundation-model side, the Segment Anything Model (SAM)~\cite{kirillov2023segment} and its successor SAM~2~\cite{ravi2024sam2}, together with medical adaptations such as MedSAM~\cite{ma2024segment} and Medical-SAM-Adapter~\cite{wu2025medical}, transfer representations from web- or domain-scale pretraining to segmentation tasks via prompting or lightweight adapters. A recent re-examination of nnU-Net~\cite{isensee2024nnunet} further argues that well-tuned U-Net variants remain difficult to beat once experimental protocols are controlled rigorously, and contemporary surveys of efficient medical segmentation in the era of Mamba and foundation models~\cite{shu2026efficient}, together with meta-learning frameworks for medical imaging~\cite{nguyen2022metalearning}, catalog the broader landscape of work that, like Active-DiNTS, targets data efficiency in the clinical setting. We see these efforts as complementary to Active-DiNTS rather than competing. They trade larger pretraining or compute budgets for stronger representations, while Active-DiNTS targets settings in which neither pretraining corpora nor multi-GPU resources are available and the architecture must be discovered from scratch under a tight annotation budget. The AL machinery introduced in this paper could, in principle, sit on top of a transformer or foundation-model backbone; we discuss that combination in Sec.~\ref{sec:future}.

\section{Methodology}
\label{sec:method}

\subsection{Search Space and One-Shot Differentiable Topology}
\label{sec:searchspace}

\begin{figure}[tbp]
  \centering
  \includegraphics[width=\linewidth]{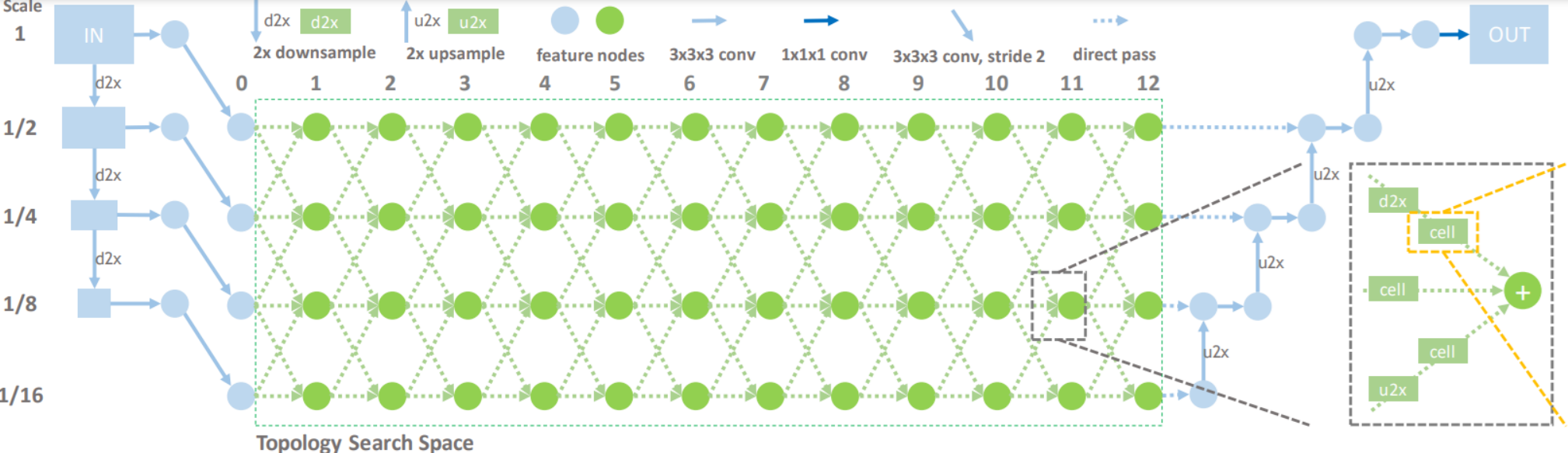}
  \caption{Topology and cell search spaces explored by Active-DiNTS. The macro graph (top) interconnects $L{=}12$ layers across $D{=}4$ resolutions; green dashed edges carry one of $O{=}5$ cell operations. Adapted from He et al.~\cite{he2021dints}.}
  \label{fig:searchspace}
\end{figure}

Active-DiNTS operates within the cell-based, multi-resolution search space inherited from DiNTS~\cite{he2021dints} and depicted in \Cref{fig:searchspace}. The space is structured at two granularities:

\begin{itemize}
  \item \textbf{Macro level (topology).} A directed graph with $L{=}12$ layers spans $D{=}4$ spatial resolutions. Adjacent resolutions are connected through fully populated edges that carry a fixed catalog of operations, namely $2\times$ downsample, $2\times$ upsample, $3{\times}3{\times}3$ convolution, $1{\times}1{\times}1$ convolution, $3{\times}3{\times}3$ strided convolution, and direct pass (skip), which together dictate how features flow from input to output.
  \item \textbf{Micro level (cell).} Every edge instantiates one of $O{=}5$ candidate operators: (0) skip, (1) full 3D $3{\times}3{\times}3$ convolution, (2) pseudo-3D P3D $3{\times}3{\times}1$, (3) P3D $3{\times}1{\times}3$, or (4) P3D $1{\times}3{\times}3$~\cite{zhu2019vnas}. With the exception of skip, all cells precede and follow their convolution with ReLU activation and Instance Normalization.
\end{itemize}

\noindent In contrast to cell-repetition spaces such as DARTS~\cite{liudarts} or Auto-DeepLab~\cite{liu2019auto}, this design lets each cell be searched independently, so the resulting encoder--decoder configurations form a substantially richer family.

\paragraph{Continuous relaxation and discretization.}
As in standard one-shot NAS, each discrete operator choice is relaxed into a continuous probability vector and trained by gradient descent. When the search ends, the operator with the highest probability on each edge is kept and the rest are pruned, yielding a deployable network. A known shortcoming of this protocol is the \emph{discretization gap}: edges whose probabilities are non-negligible but lie just below the cutoff are dropped, perturbing the feature flow the search converged to. DiNTS attenuates this issue with a topology loss that pushes high-probability connections into feasible paths, a mechanism that Active-DiNTS preserves. A RAM-cost factor $\sigma$ controls the density of the final graph; larger $\sigma$ produces denser, more interconnected topologies at higher GPU-memory cost. All experiments in this paper use $\sigma{=}0.8$, the most resource-heavy but most accurate setting.

\subsection{Active Learning with Uncertainty-Driven Querying}
\label{sec:al}

The architecture search in Active-DiNTS runs inside a pool-based AL loop. Two distinct unlabeled pools are maintained in parallel:

\begin{itemize}
  \item \textbf{Outer pool $U_W$ (weight updates).} At every query round the current model produces segmentation masks for all volumes in $U_W$. An uncertainty heuristic scores each prediction; the $N_Q$ volumes with the highest scores are dispatched to the oracle, a domain expert in deployment, or the ground-truth labels in our simulation. Once annotated, these volumes migrate from $U_W$ into the labeled set $L_W$, which feeds $T$ epochs of weight optimization.
  \item \textbf{Inner pool $U_T$ (topology updates).} After the weight step, the model predicts on $U_T$. The single most uncertain volume is labeled and added to $L_T$, from which one mini-batch is drawn to update the topology parameters under DiNTS's bi-level optimization. Because $U_T$ is drained one volume per round while $U_W$ is drained $N_Q$ at a time, $U_T$ is refilled from the original data whenever it empties before $U_W$ does.
\end{itemize}

Three AL regimes determine how uncertainty-driven selection is applied to each loop:

\begin{enumerate}
  \item \emph{Weights-only}: uncertainty drives $U_W$ selection; $U_T$ samples are picked at random.
  \item \emph{Topology-only}: uncertainty drives $U_T$ selection; $U_W$ samples are picked at random.
  \item \emph{Joint (weights \& topology)}: uncertainty drives both pools simultaneously.
\end{enumerate}

\paragraph{Uncertainty measures.}
We rely on three lightweight, widely used heuristics. Given a predicted segmentation mask $D$ whose per-voxel label distribution over $y \in \mathcal{Y}$ is $P(y)$, the \emph{Entropy} of $D$ is
\begin{align}
  H(D) &= -\sum_{y\in\mathcal{Y}} P(y)\,\log_2 P(y). \label{eq:entropy}
\end{align}
For a mask seen as an array of $N$ values $\{x_i\}_{i=1}^N$ (softmax probabilities or logits, depending on configuration), we also compute the \emph{Variance}
\begin{align}
  \sigma^2 &= \frac{1}{N}\sum_{i=1}^{N}(x_i-\mu)^2, \label{eq:var}
\end{align}
and the \emph{Standard Deviation}
\begin{align}
  \sigma &= \sqrt{\frac{1}{N}\sum_{i=1}^{N}(x_i-\mu)^2}, \label{eq:std}
\end{align}
with $\mu$ the mean of $\{x_i\}$.

Entropy registers the overall disorder of the predicted distribution and is therefore well suited to flag volumes on which the model is undecided across classes. Variance and Standard Deviation, in contrast, capture how widely individual voxel predictions deviate from their mean, which is useful when the model is over-confident on a dominant class but uncertain on smaller subregions. Standard Deviation is the more sensitive of the two to extreme values, which makes it a reasonable signal for surfacing atypical volumes but also exposes it to degenerate cases such as scans with severe imaging artifacts, where uniformly high voxel-wise uncertainty carries little information about the tumor itself.

\subsection{Bi-level Training with Coupled Pools}
\label{sec:algo}

\begin{figure}[tbp]
  \centering
  \includegraphics[width=\linewidth]{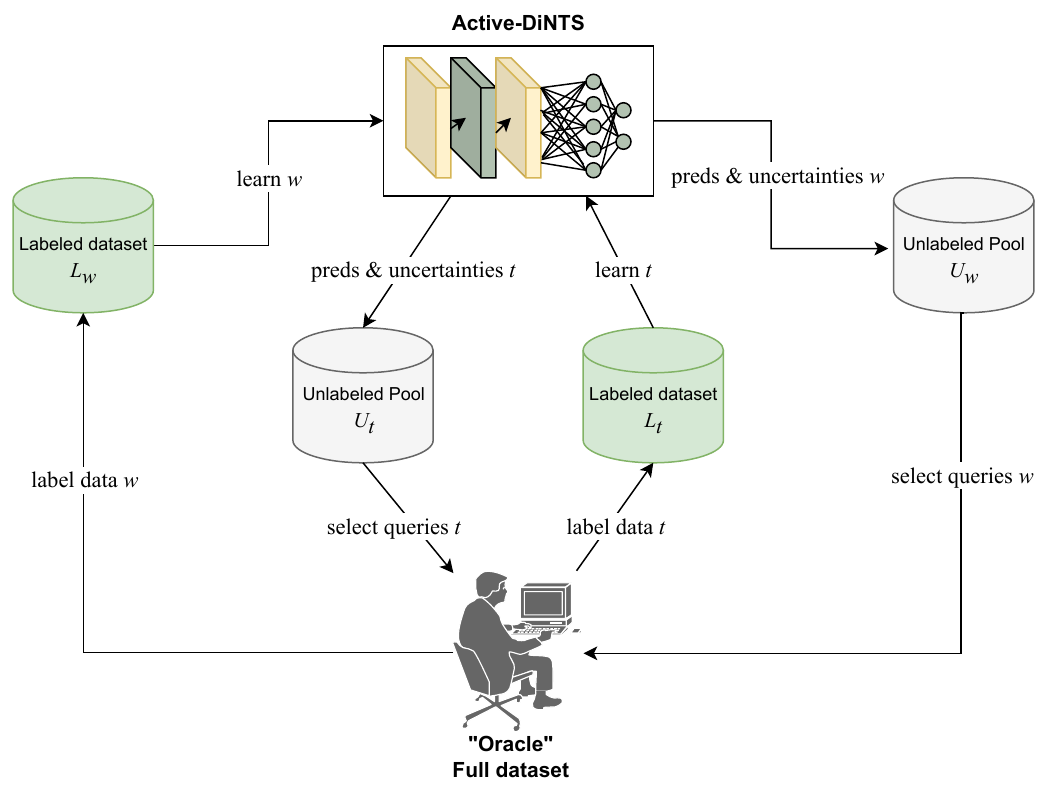}
  \caption{Active-DiNTS workflow. (1)~Unlabeled pools $U_W$ and $U_T$ are scored by uncertainty; (2)~the oracle annotates the most informative volumes; (3)~bi-level optimization alternates between weights and topology.}
  \label{fig:workflow}
\end{figure}

\Cref{fig:workflow} gives a high-level picture of the pipeline; \Cref{alg:activedints} renders it formally. Each outer iteration corresponds to a single AL query round. Once $U_W$ is depleted, the continuous architecture is discretized into a deployable model. A key point is that, because the AL loop has been training with steadily more informative data all along, the searched model needs \emph{no} separate retraining phase. The architecture delivered at the end of the search is already the final model. This stands in contrast to DiNTS, which reinitializes and retrains the discretized architecture with double the filter count and batch size.

\begin{algorithm}[H]
\DontPrintSemicolon
\SetKwInOut{Input}{Input}\SetKwInOut{Output}{Output}
\Input{Search space $\mathcal{S}$; model $M$; oracle dataset $O$; labeled sets $L_W,L_T \leftarrow \emptyset$; unlabeled pools $U_W,U_T$; query strategy $QS$; query function $QF$; samples per query $N_Q$; training epochs $T$; topology batch size $B$.}
\Output{Trained, discretized Active-DiNTS $M$.}
$M \leftarrow \texttt{initialize}(\mathcal{S})$ \;
\While{$U_W \neq \emptyset$}{
  $\texttt{preds} \leftarrow \texttt{predict}(M, U_W)$ \;
  \If{$QS$ is \texttt{topology-only}}{$QF \leftarrow$ \texttt{RandomSelection}}
  $\texttt{selected} \leftarrow \texttt{select}(\texttt{preds}, QF, N_Q)$ \;
  $\texttt{labeled} \leftarrow \texttt{label}(\texttt{selected}, O)$ \;
  $L_W, U_W \leftarrow \texttt{update\_pools}(\texttt{labeled})$ \;
  \For{$t \in \{1,\dots,T\}$}{$\texttt{train\_weights}(M, L_W)$}
  \If{$U_T$ is empty}{$U_T \leftarrow \texttt{replenish}()$}
  $\texttt{preds}_T \leftarrow \texttt{predict}(M, U_T)$ \;
  \If{$QS$ is \texttt{weights-only}}{$QF \leftarrow$ \texttt{RandomSelection}}
  $\texttt{selected}_T \leftarrow \texttt{select}(\texttt{preds}_T, QF, 1)$ \;
  $\texttt{labeled}_T \leftarrow \texttt{label}(\texttt{selected}_T, O)$ \;
  $L_T, U_T \leftarrow \texttt{update\_pools}(\texttt{labeled}_T)$ \;
  $\texttt{batch} \leftarrow \texttt{sample\_batch}(L_T, B)$ \;
  $\texttt{train\_topology}(M, \texttt{batch})$ \;
}
\caption{Active-DiNTS: bi-level topology search with pool-based AL.}
\label{alg:activedints}
\end{algorithm}

The bi-level structure follows the conventions of differentiable NAS~\cite{liudarts,liu2019auto,he2021dints} in that weights and architecture parameters are optimized on separate data splits. What is new here is that both splits are not fixed partitions of a fully labeled dataset, but rather grow incrementally under the control of the AL loop. As the search progresses, the weight-update step becomes heavier because $L_W$ accumulates samples, while the prediction-and-scoring step becomes lighter because $U_W$ shrinks; the per-round cost is therefore approximately constant.

The topology update relies on the composite loss originally proposed in DiNTS: (i)~a segmentation term combining Dice and cross-entropy, (ii)~a topology term that penalizes infeasible edges, (iii)~an entropy regularizer that drives architecture probabilities toward binary values, and (iv)~a RAM-cost term that caps peak memory. Weights are optimized with SGD at 0.9 momentum, $4{\times}10^{-5}$ weight decay, and a cosine-annealing schedule, whereas topology parameters use Adam.

\section{Experimental Setup}
\label{sec:setup}

\subsection{Dataset}

Active-DiNTS is evaluated on \textbf{MSD Task01 BrainTumour}~\cite{antonelli2022medical}, the publicly released subset of the Medical Segmentation Decathlon that derives from the BraTS challenge~\cite{menze2014multimodal}. Its training partition contains 484 multi-parametric MRI volumes, each one bringing four modalities, namely Fluid-Attenuated Inversion Recovery (FLAIR), T1-weighted (T1w), post-Gadolinium contrast T1-weighted (T1w-Gd), and T2-weighted (T2w), together with three tumor-subregion labels, Edema, Enhancing tumor, and Non-Enhancing core. The cases come from patients with glioblastoma or lower-grade glioma. Following He et al.~\cite{he2021dints}, the 484 volumes are partitioned into 388 for training and 96 for testing under 5-fold cross-validation. Within each training fold, half of the examples populate the weight-update pool $U_W$ and the other half populate the topology-update pool $U_T$. Task01 BrainTumour is used as the primary benchmark for three reasons: (i)~it is the most studied task in MSD and admits direct, like-for-like comparisons with every NAS baseline considered here; (ii)~its multi-modal MRI input and three-class subregion structure stress the topology search across heterogeneous feature distributions; and (iii)~the published DiNTS results on this task provide an especially strong reference point against which to measure improvements.

\subsection{Baselines}

We compare Active-DiNTS against two families of methods. The \textbf{hand-designed} group covers CerebriuDIKU (Multi-Planar U-Net)~\cite{perslev2019one}, nnU-Net~\cite{isensee2019nnunet}, 3D U-Net~\cite{cciccek20163d}, Attention U-Net~\cite{oktay2018attention}, and NVDLMED (UMCT)~\cite{xia20203d}. The \textbf{NAS} group covers SCNAS~\cite{kim2019scalable}, C2FNAS~\cite{yu2020c2fnas}, and DiNTS~\cite{he2021dints}. For DiNTS specifically we report two variants: DiNTS (Pancreas), whose architecture was searched on the MSD Pancreas task and used as the headline comparison in the original paper, and DiNTS (Brain), whose architecture was searched on BrainTumour itself and therefore acts as the more demanding, task-specific reference. All baseline numbers are reproduced from the original publications rather than re-run here, which means the comparison inherits whatever differences in preprocessing and post-processing those papers used beyond the shared 5-fold split.

\subsection{Active-DiNTS Configurations}

Each Active-DiNTS training run is parameterized by a query function $\in \{\texttt{Random}, \texttt{Entropy}, \texttt{Variance}, \texttt{Std}\}$, an Active Learning setup strategy $\in \{\texttt{weights-only}, \texttt{topology-only}, \texttt{joint}\}$, and a per-query budget $N_Q \in [2, 100]$. The training epochs allocated to each query lie in $[1, 1430]$, of which a warm-up may consume up to 50\%. These ranges were explored by grid search under the same 5-fold cross-validation used for evaluation, with the epoch and warm-up bounds taken from the DiNTS setup~\cite{he2021dints}. The four MRI modalities map to the four input channels of the network. When computing uncertainty, the query axis can be set to operate over modalities (Axis~0) or over slices (Axis~3); this controls the dimension along which dispersion is measured. The Random query function plays the role of a lower-bound control throughout the experiments.

\subsection{Evaluation Metrics and Protocol}

Segmentation accuracy is reported with the Dice--S\o rensen Coefficient (DSC), $\mathrm{DSC} = 2|X \cap Y| / (|X| + |Y|)$, with $X$ and $Y$ the predicted and ground-truth voxel sets, respectively. We report the DSC of each tumor subregion, DSC1 for Edema, DSC2 for Enhancing, and DSC3 for Non-Enhancing, together with the unweighted mean of the three. Computational profiles are summarized by FLOPs (GigaFLOPs), trainable parameters (millions), peak GPU memory during search (MB), number of GPUs, and wall-clock search time expressed in GPU-days.

\subsection{Implementation Details}

Active-DiNTS is implemented in Python on top of PyTorch~\cite{NEURIPS2019_9015}, with MONAI~\cite{cardoso2022monai} providing medical-imaging preprocessing, data augmentation, and the Dice + cross-entropy loss. SciPy~\cite{2020SciPy-NMeth} supplies the numerical primitives used by the uncertainty heuristics. All searches and evaluations are run on a single Linux workstation with an Intel Xeon Silver 4215 CPU (2.50\,GHz, 8 cores), 126\,GB of RAM, and one NVIDIA Tesla V100 GPU with 32\,GB of VRAM. Unlike DiNTS, which searches on one dataset and then reinitializes and retrains the resulting architecture with double the capacity, Active-DiNTS deploys the searched model as-is, since the AL-guided search already produces a model of sufficient quality. Multi-channel preprocessing and augmentations follow the standard BraTS/DiNTS protocol. Architecture parameters are initialized from a Gaussian distribution; weight-update SGD uses 0.9 momentum and $4{\times}10^{-5}$ weight decay with cosine annealing, while topology parameters use Adam. The RAM factor is held fixed at $\sigma{=}0.8$. The complete implementation is publicly available.\footnote{\url{https://github.com/geantrindade/Active-DiNTS}}

\section{Results and Discussion}
\label{sec:results}

\subsection{Learning Curves and Active Learning Dynamics}
\label{sec:curves}

\begin{figure}[tbp]
  \centering
  \includegraphics[width=\linewidth]{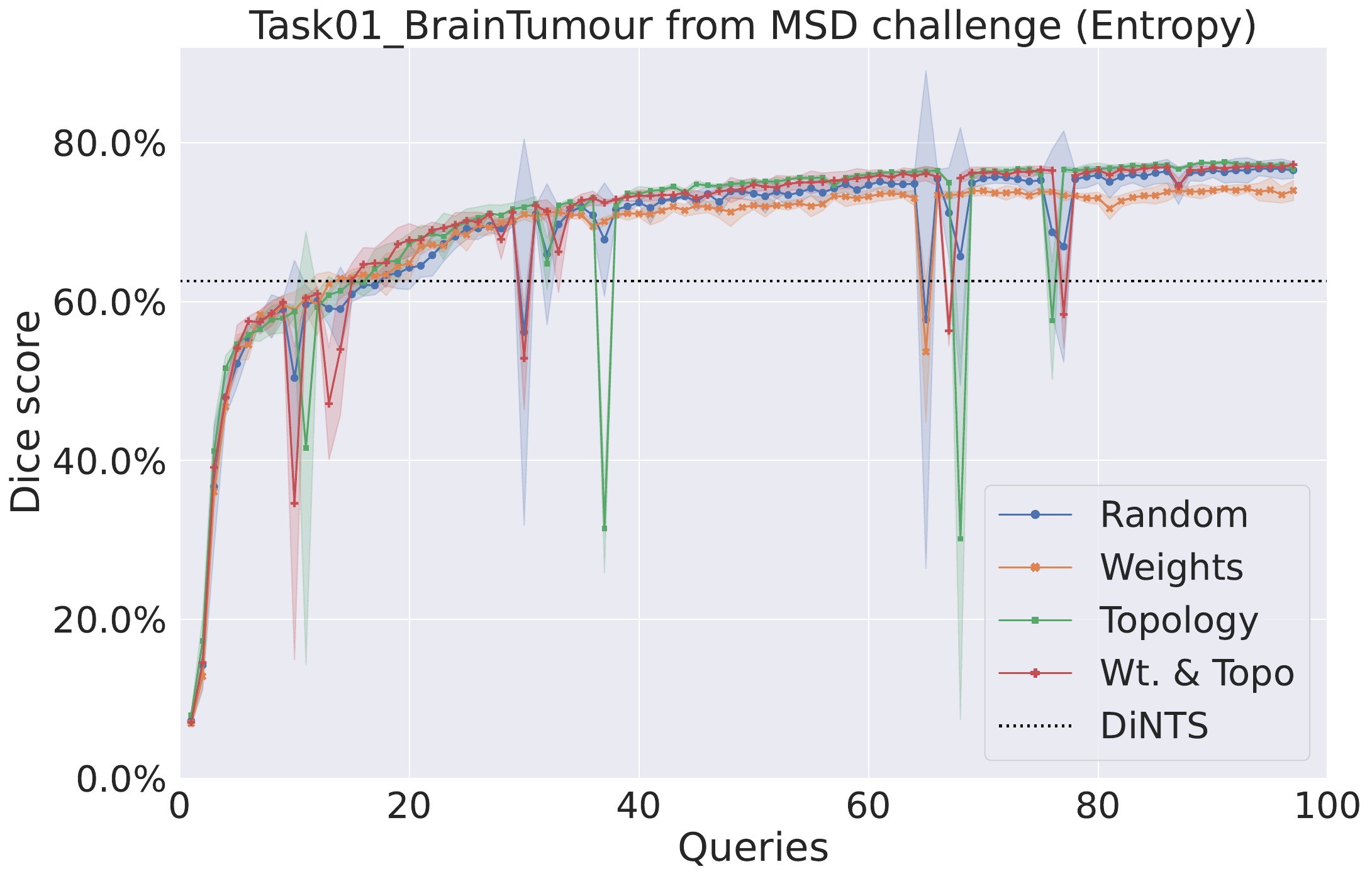}
  \caption{Learning curves for the joint AL setup. Uncertainty-driven strategies (Entropy, Variance, Std) are contrasted with Random selection and the DiNTS baseline (horizontal line).}
  \label{fig:curves-uncertainty}
\end{figure}

\Cref{fig:curves-uncertainty} reports Dice as a function of the number of AL queries for the \emph{joint} (weights \& topology) configuration under each uncertainty heuristic, with the Random baseline and the DiNTS reference plotted for context.

A few observations are worth highlighting. First, every Active-DiNTS variant surpasses the DiNTS baseline after roughly 18 queries, a point at which fewer than 20\% of the training volumes have been labeled. This indicates that targeted sample selection reaches competitive accuracy considerably faster than training on a fixed, fully labeled corpus. Second, uncertainty-driven heuristics outperform Random querying consistently; Entropy and Variance hold a slight advantage in the early stage (queries~1--20), while Std edges ahead afterward. Third, Dice gains are concentrated in the first 20 queries, where each newly annotated volume carries the most information relative to the still-small labeled pool; beyond about 60\% data usage, improvements are incremental. The occasional dip at individual queries suggests that certain volumes briefly shift the training data distribution, and that the short training window between consecutive queries is sometimes insufficient to absorb that shift, though recovery is fast. We did not run a controlled experiment isolating which volume properties cause these dips, so this reading of them remains a hypothesis.

\begin{figure}[tbp]
  \centering
  \includegraphics[width=\linewidth]{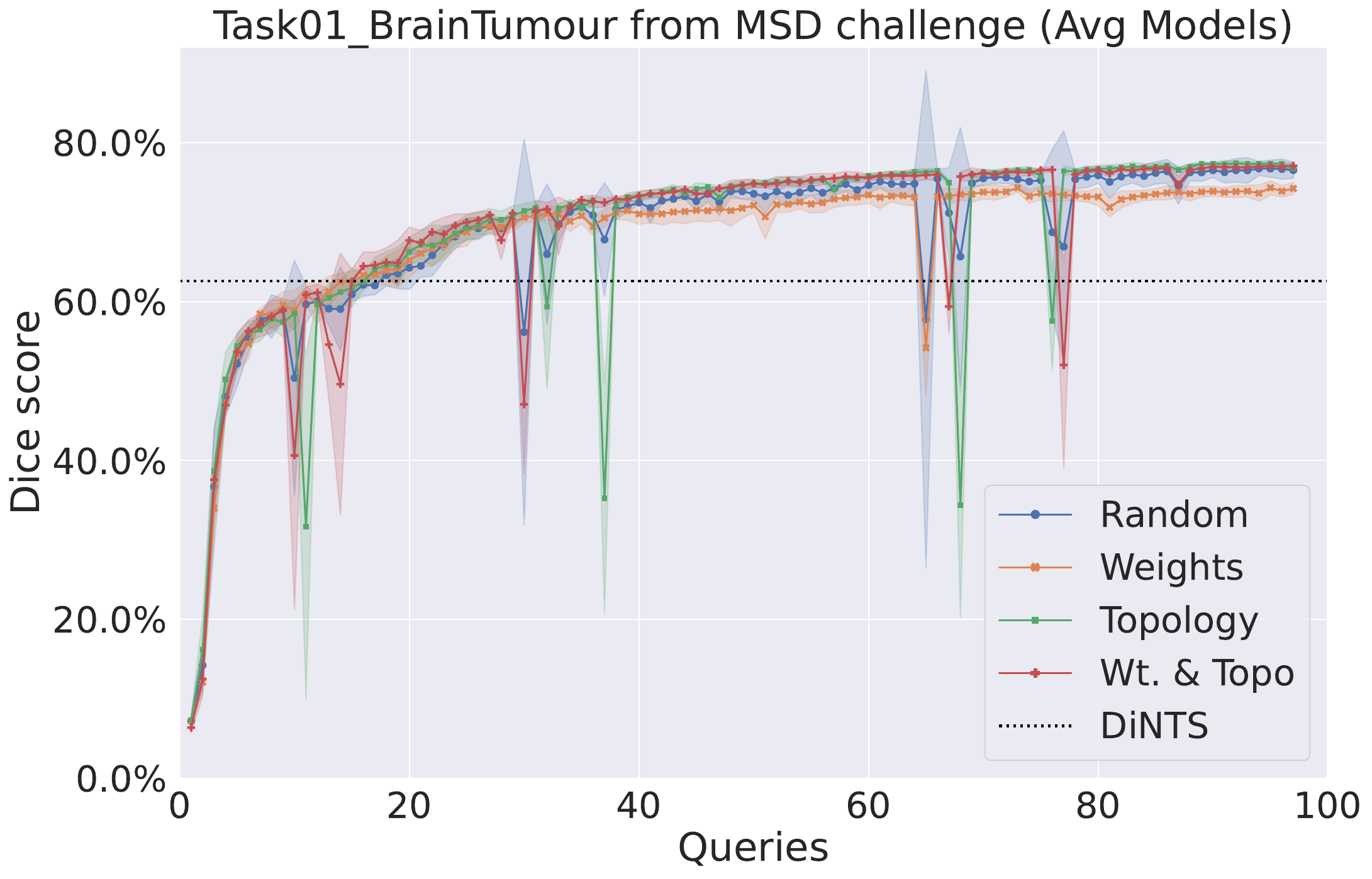}
  \caption{Dice curves averaged over uncertainty heuristics for each AL setup (weights-only, topology-only, joint), shown against the Random and DiNTS baselines.}
  \label{fig:curves-setups}
\end{figure}

\Cref{fig:curves-setups} contrasts AL setups by averaging across the three uncertainty functions. The joint configuration (weights \& topology) leads throughout the run, which is consistent with directing informative data toward both optimization targets at the same time. The topology-only setup tracks the joint curve closely, suggesting that actively refining the architecture is itself a substantial source of value, even when the weights see only randomly chosen samples. The weights-only setup is competitive until around query~35 and then plateaus, which is consistent with the topology becoming the binding constraint once the weights are well calibrated. We note that the Random curve, although inferior to weights-only past the mid-point, carries a considerably larger standard deviation across runs, so that particular ordering should be read with caution. All three setups reach state-of-the-art performance using roughly 20\% of the original annotation budget.

\subsection{Discovered Architectures}
\label{sec:arches}

\Cref{tab:ops} reports the operation distributions of the best-performing architecture discovered by each uncertainty heuristic, side by side with the reference DiNTS topology (searched on Pancreas). All Active-DiNTS topologies are fully connected (120/120 edges), whereas DiNTS retains only 97/120 edges, evidence that the AL-driven search settles on denser graphs and therefore preserves more information pathways.

\begin{table}[tbp]
  \centering
  \caption{Operation statistics for the best Active-DiNTS architectures versus DiNTS.}
  \label{tab:ops}
  \resizebox{\linewidth}{!}{%
  \begin{tabular}{lccc}
    \toprule
    Method & Most frequent op & Least frequent op & Total ops \\
    \midrule
    DiNTS & 3D conv $3{\times}3{\times}3$ (34.0\%) & P3D $3{\times}1{\times}3$ (12.4\%) & 97/120 \\
    Active-DiNTS (Random) & 3D conv $3{\times}3{\times}3$ (26.7\%) & skip (15.0\%) & 120/120 \\
    Active-DiNTS (Entropy) & 3D conv $3{\times}3{\times}3$ (27.5\%) & P3D $1{\times}3{\times}3$ (15.0\%) & 120/120 \\
    Active-DiNTS (Variance) & 3D conv $3{\times}3{\times}3$ (29.2\%) & P3D $3{\times}3{\times}1$ (13.3\%) & 120/120 \\
    Active-DiNTS (Std) & 3D conv $3{\times}3{\times}3$ (26.6\%) & skip (15.8\%) & 120/120 \\
    \bottomrule
  \end{tabular}}
\end{table}

Across all strategies, the standard $3{\times}3{\times}3$ convolution remains the dominant operation, in line with its broad receptive field being well matched to volumetric MRI features. What is striking is that Active-DiNTS distributes operations more evenly than DiNTS. The spread between the most and the least frequent operation shrinks from roughly 22 percentage points in DiNTS to between 11 and 16 points in Active-DiNTS. This is consistent with the incremental, uncertainty-guided training pushing the search to exploit a wider mix of convolution patterns, although with a single searched topology per strategy we cannot separate that effect from run-to-run variance. The P3D variants, which factorize the 3D kernel along different spatial axes, complement full 3D convolutions by providing anisotropic feature extraction at a lower parameter cost. Skip connections appear comparatively less often in Active-DiNTS topologies, which may reflect the denser connectivity already providing ample gradient pathways.

\begin{sidewaysfigure}
  \centering
  \includegraphics[width=\textheight]{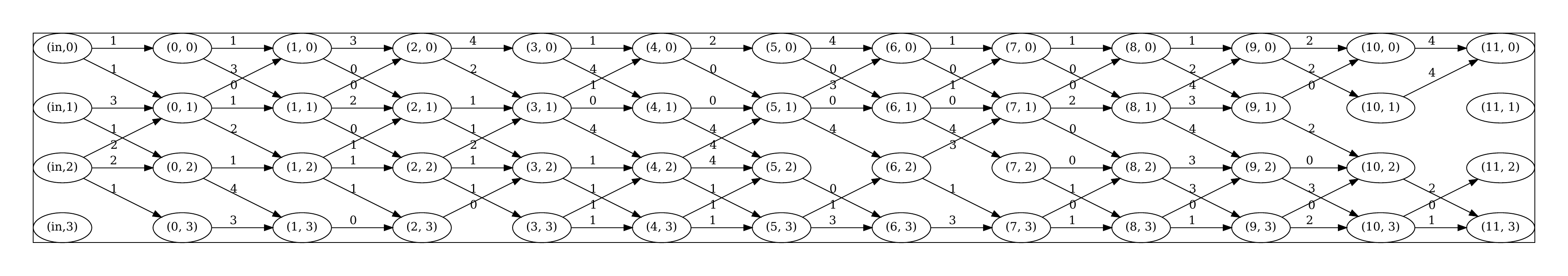}
  \caption{Reference DiNTS architecture searched on the Pancreas dataset. Edge labels: 0\,=\,Skip, 1\,=\,$3{\times}3{\times}3$, 2\,=\,P3D $3{\times}3{\times}1$, 3\,=\,P3D $3{\times}1{\times}3$, 4\,=\,P3D $1{\times}3{\times}3$.}
  \label{fig:dints-graph}
\end{sidewaysfigure}

\begin{sidewaysfigure}
  \centering
  \subfloat[Random (Query 41/97).]{\includegraphics[width=\textheight]{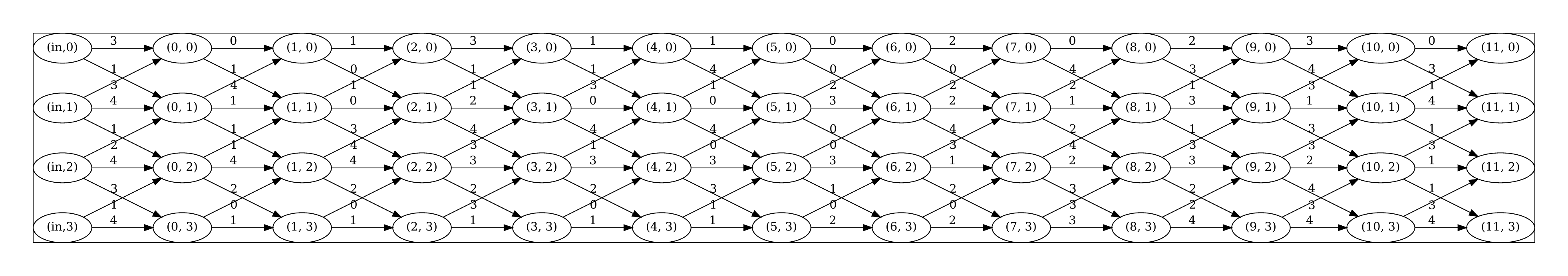}}\\
  \subfloat[Entropy (Query 35/97).]{\includegraphics[width=\textheight]{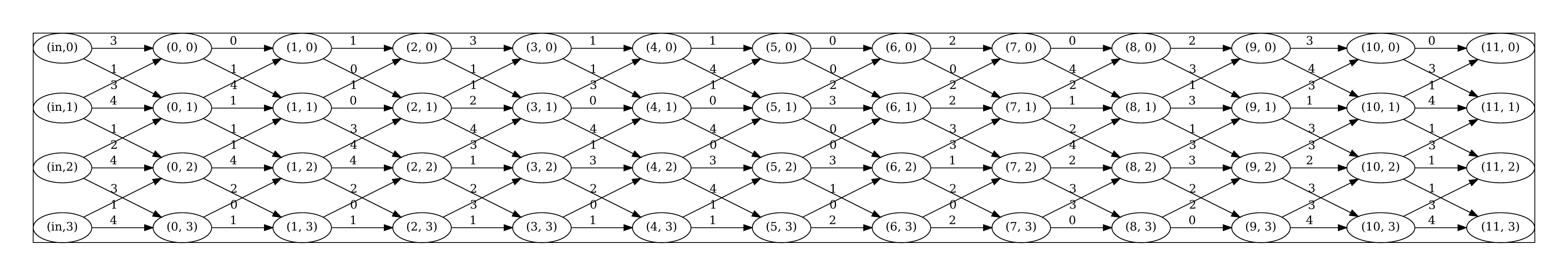}}\\
  \subfloat[Variance (Query 36/97).]{\includegraphics[width=\textheight]{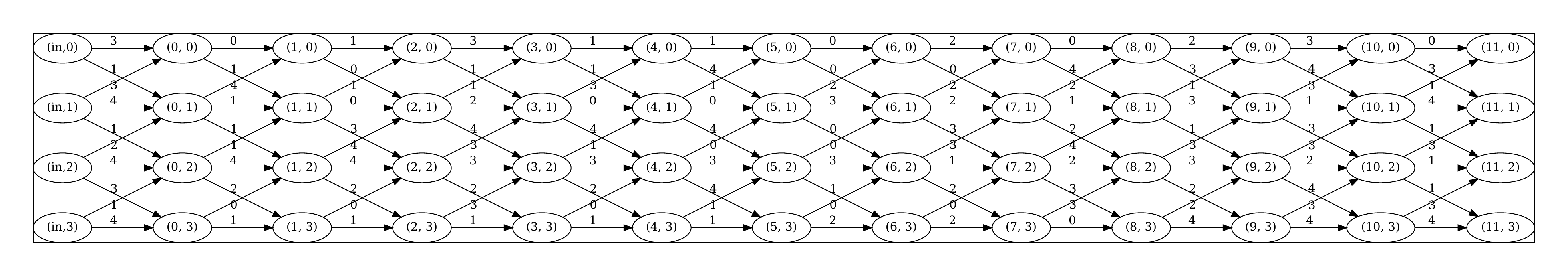}}\\
  \subfloat[Std (Query 37/97).]{\includegraphics[width=\textheight]{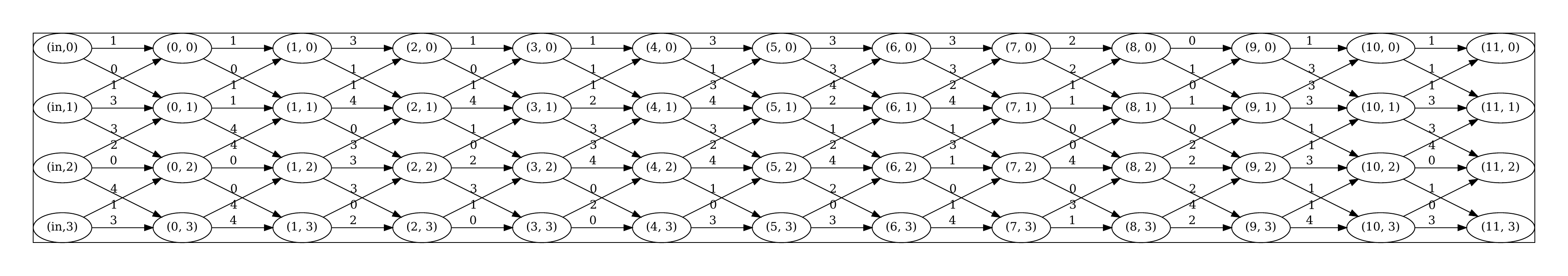}}
  \caption{Active-DiNTS architectures (Random / Entropy / Variance / Std) at the earliest query that surpasses DiNTS (Brain) on BrainTumour. Edge coding follows \Cref{fig:dints-graph}.}
  \label{fig:active-graphs}
\end{sidewaysfigure}

\Cref{fig:dints-graph,fig:active-graphs} display the reference DiNTS graph and the four Active-DiNTS graphs, respectively.

\subsection{Quantitative Comparison on MSD/Task01 BrainTumour}
\label{sec:quant}

\paragraph{Dice per sub-region.}
\Cref{tab:dice} compares average Dice scores across baselines and Active-DiNTS variants. For each uncertainty heuristic we list three checkpoints: the best final model at the end of the full query budget~(*), the earliest checkpoint that exceeds DiNTS (Pancreas) in average Dice~($\dagger$), and the earliest checkpoint that exceeds DiNTS (Brain)~($\ddagger$).

\begin{table}[tbp]
\centering
\caption{Average Dice--S\o rensen Coefficient (DSC) on Task01 BrainTumour. Sub\-regions: Edema (DSC1), Enhancing tumor (DSC2), Non-Enhancing core (DSC3).}
\label{tab:dice}
\begin{tabular}{lcccc}
\toprule
Method & DSC1 & DSC2 & DSC3 & Average \\
\midrule
CerebriuDIKU~\cite{perslev2019one} & 69.52 & 43.11 & 66.74 & 59.79 \\
NVDLMED~\cite{xia20203d} & 67.52 & 45.00 & 68.01 & 60.18 \\
SCNAS~\cite{kim2019scalable} & 67.40 & 45.75 & 68.26 & 60.47 \\
nnU-Net~\cite{isensee2019nnunet} & 68.04 & 46.81 & 68.46 & 61.10 \\
C2FNAS~\cite{yu2020c2fnas} & 67.62 & 48.60 & 69.72 & 61.98 \\
DiNTS (Pancreas)~\cite{he2021dints} & 69.28 & 48.65 & 69.75 & 62.56 \\
DiNTS (Brain)~\cite{he2021dints} & 80.20 & 61.09 & 77.63 & 72.97 \\
\midrule
\textbf{Active-DiNTS (Entropy)}* & \textbf{90.65} & 60.04 & \textbf{82.81} & \textbf{77.83} \\
Active-DiNTS (Entropy)${}^\dagger$ & 84.58 & 38.71 & 67.67 & 63.65 \\
Active-DiNTS (Entropy)${}^\ddagger$ & 87.45 & 54.24 & 78.86 & 73.52 \\
\textbf{Active-DiNTS (Variance)}* & 90.11 & 59.58 & 82.50 & 77.40 \\
Active-DiNTS (Variance)${}^\dagger$ & 81.96 & 39.64 & 69.70 & 63.77 \\
Active-DiNTS (Variance)${}^\ddagger$ & 88.19 & 53.25 & 79.25 & 73.56 \\
\textbf{Active-DiNTS (Std)}* & 90.27 & \textbf{60.14} & \textbf{82.89} & \textbf{77.77} \\
Active-DiNTS (Std)${}^\dagger$ & 83.16 & 36.69 & 69.32 & 63.06 \\
Active-DiNTS (Std)${}^\ddagger$ & 87.80 & 54.42 & 78.85 & 73.69 \\
\bottomrule
\end{tabular}

\vspace{0.4em}
\footnotesize * Best final performance \quad
${}^\dagger$ Overcomes DiNTS (Pancreas) \quad
${}^\ddagger$ Overcomes DiNTS (Brain)
\end{table}

The headline result is that the three best Active-DiNTS models (Entropy*, Variance*, Std*) all clear the strongest baseline, DiNTS (Brain), by about 5 points of average Dice. The gains concentrate on Edema and Non-Enhancing, which improve by roughly 10 and 5 percentage points respectively, while Enhancing tumor performance stays within a single point of the best baseline. Enhancing is the hardest of the three subregions due to its small spatial extent and irregular shape; that all methods cluster around the same score on this class suggests it may be more responsive to post-processing or loss-function changes than to topological ones, though we did not test that directly.

Among the intermediate checkpoints, the $\dagger$ variants already beat C2FNAS, nnU-Net, and DiNTS (Pancreas) after consuming fewer than 15\% of the available labels. The $\ddagger$ variants overtake DiNTS (Brain) at roughly 35--40 queries, i.e.\ around 35--40\% of the total budget. These intermediate results matter in practice. They describe what Active-DiNTS does when annotation is strictly capped.

\paragraph{Computational resources.}
\Cref{tab:compute} juxtaposes the resource profiles of all methods.

\begin{table}[tbp]
\centering
\caption{Compute comparison: FLOPs (GFLOPs), trainable parameters (M), peak GPU memory (MB), number of GPUs, search time (GPU-days), and average Dice.}
\label{tab:compute}
\resizebox{\linewidth}{!}{%
\begin{tabular}{lcccccc}
\toprule
Method & FLOPs & Params & Memory & GPU(s) & Time & Avg Dice \\
\midrule
CerebriuDIKU & -- & 62 & -- & -- & -- & 59.79 \\
3D U-Net & 658 & 18 & 9176 & 1 (TitanX) & 3 & -- \\
Attention U-Net & 1163 & 104 & 13465 & -- & -- & -- \\
SCNAS & 425 & -- & -- & 4 (V100) & 1 & 60.47 \\
C2FNAS & 151 & 17 & 5730 & 8 (V100) & 333 & 61.98 \\
DiNTS (Pancreas) & 334 & 152 & 13018 & 8 (V100) & 5.8 & 62.56 \\
DiNTS (Brain) & 334 & 152 & 13018 & 8 (V100) & 5.8 & 72.97 \\
\midrule
\textbf{Active-DiNTS (Entropy)}* & 8500 & 127 & 17249 & 1 (V100) & 5.5 & 77.83 \\
Active-DiNTS (Entropy)${}^\dagger$ & 8690 & 123 & 14802 & 1 (V100) & 0.33 & 63.65 \\
Active-DiNTS (Entropy)${}^\ddagger$ & 8770 & 152 & 15483 & 1 (V100) & 1.1 & 73.52 \\
\textbf{Active-DiNTS (Variance)}* & 8360 & 153 & 17283 & 1 (V100) & 6.4 & 77.40 \\
Active-DiNTS (Variance)${}^\dagger$ & 8720 & 125 & 14835 & 1 (V100) & 0.21 & 63.77 \\
Active-DiNTS (Variance)${}^\ddagger$ & 9650 & 126 & 15577 & 1 (V100) & 1.4 & 73.56 \\
\textbf{Active-DiNTS (Std)}* & 8500 & 127 & 17307 & 1 (V100) & 4.6 & 77.77 \\
Active-DiNTS (Std)${}^\dagger$ & 8220 & 124 & 14736 & 1 (V100) & 0.42 & 63.06 \\
Active-DiNTS (Std)${}^\ddagger$ & 8750 & 154 & 15545 & 1 (V100) & 1.5 & 73.69 \\
\bottomrule
\end{tabular}}

\vspace{0.4em}
\footnotesize * Best final performance \quad
${}^\dagger$ Overcomes DiNTS (Pancreas) \quad
${}^\ddagger$ Overcomes DiNTS (Brain)
\end{table}

Active-DiNTS architectures are markedly denser than those of DiNTS and C2FNAS, which translates into noticeably higher FLOPs (about 8200--9650 GFLOPs against 151--658 GFLOPs). This follows directly from running with fully connected topologies at $\sigma{=}0.8$ and from the fact that the AL-driven search optimizes for predictive accuracy rather than computational economy. That said, the trainable-parameter count stays in the same range, 127 to 154\,M for Active-DiNTS versus 152\,M for DiNTS. Several intermediate Active-DiNTS configurations are actually lighter (123--126\,M).

Peak GPU memory is higher in absolute terms (14.7 to 17.3\,GB on a single 32\,GB V100), but it still represents only 46--54\% of the available VRAM, whereas DiNTS occupies roughly 13\,GB on each of eight 16\,GB V100s, i.e.\ around 81\% of each card. The total hardware bill is therefore drastically smaller for Active-DiNTS: one GPU instead of eight.

The most prominent efficiency gain shows up in search time. The fastest Active-DiNTS variant, Variance${}^\dagger$, reaches the accuracy of DiNTS (Pancreas) in just 0.21 GPU-days, more than 27$\times$ faster than the 5.8 GPU-days that DiNTS needs. Even to overtake DiNTS (Brain), the Entropy${}^\ddagger$ variant only needs 1.1 GPU-days, roughly a 5$\times$ speedup. Two factors plausibly account for these savings. The AL loop trims the effective training set down to its most informative core, and the absence of a separate retraining phase removes a major source of latency. We did not ablate these two contributions separately, so their relative weight is not established here.

For scenarios in which inference-time FLOPs are the decisive cost, for example mobile or edge deployment, post-hoc compression methods such as structured pruning and quantization could be applied to the dense Active-DiNTS models; we did not evaluate how much accuracy such compression would cost. In the medical-imaging context, however, lightweight inference is usually less critical than segmentation accuracy and efficient use of the limited annotation budget, both of which Active-DiNTS optimizes for directly.

\subsection{Qualitative Analysis}
\label{sec:qualitative}

\begin{figure}[tbp]
  \centering
  \subfloat[Input modalities (FLAIR, T1w, T1w-Gd, T2w).]{\includegraphics[width=\linewidth]{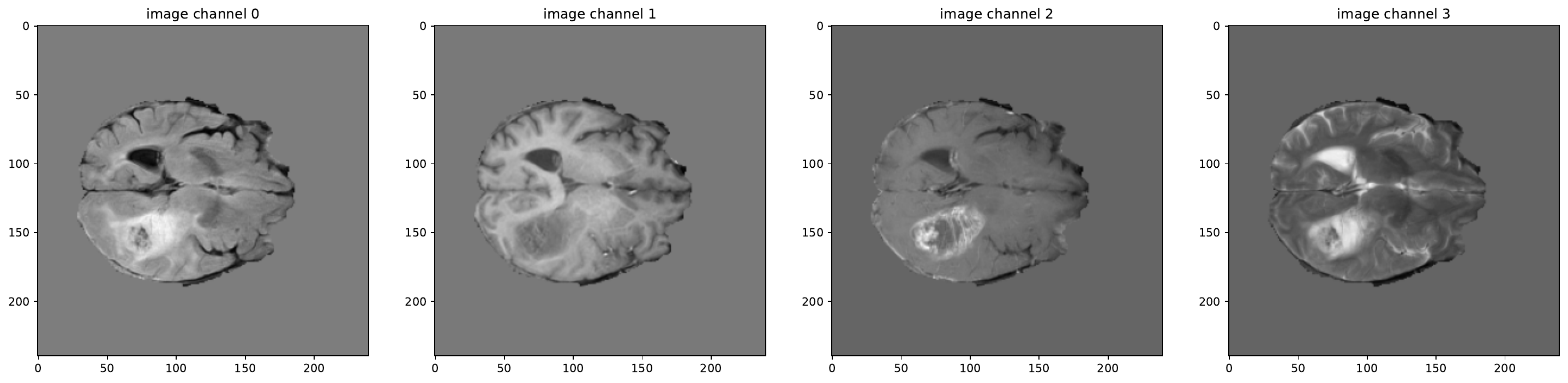}}\\[2pt]
  \subfloat[Ground-truth masks (Non-Enhancing, Edema, Enhancing).]{\includegraphics[width=0.75\linewidth]{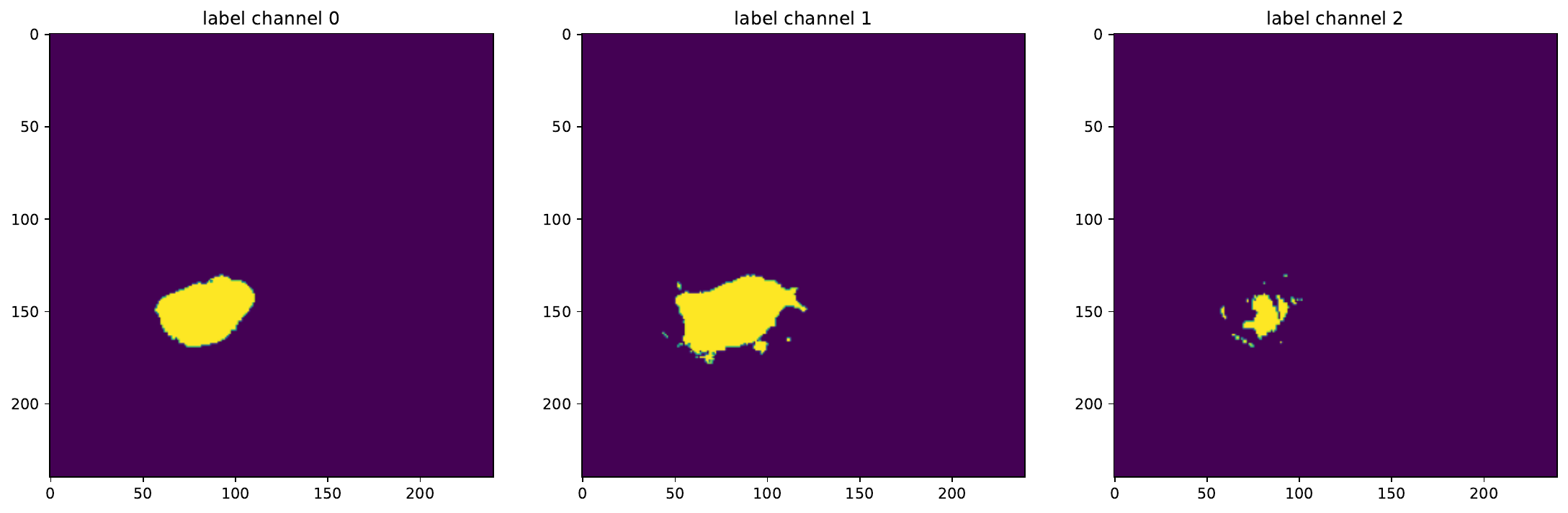}}\\[2pt]
  \subfloat[Active-DiNTS predictions.]{\includegraphics[width=0.75\linewidth]{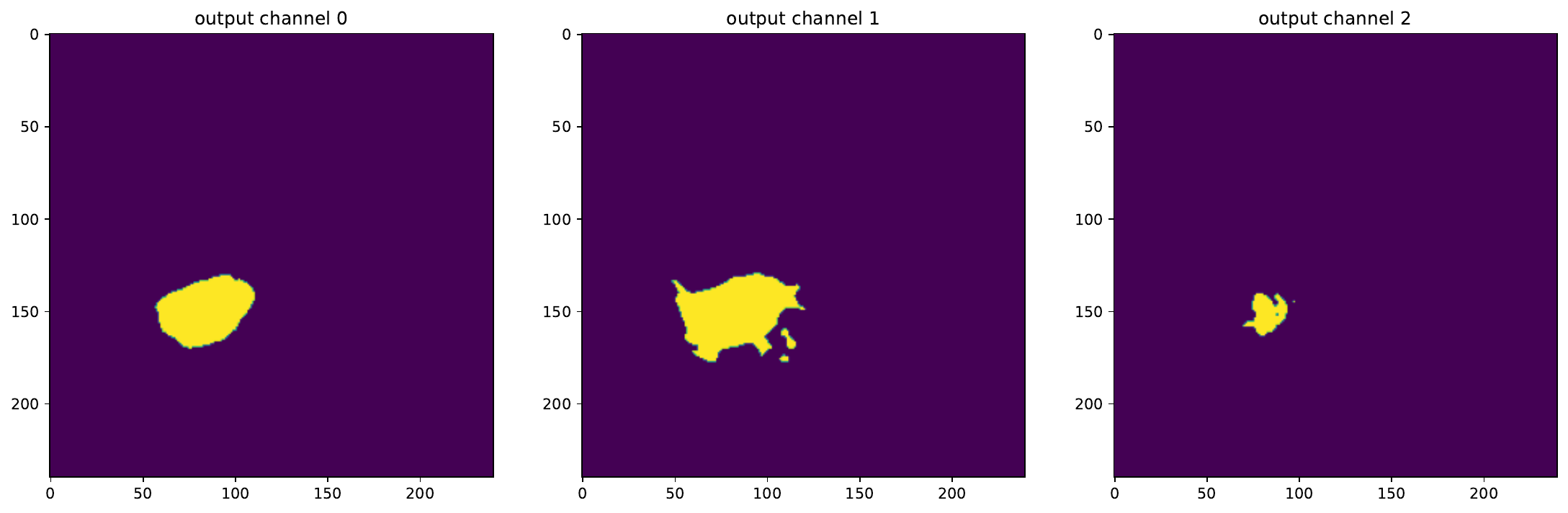}}
  \caption{Qualitative example (slice 70/150). (a)~Four MRI modalities. (b)~Ground-truth tumor sub-region masks. (c)~Predicted segmentation masks.}
  \label{fig:qualitative}
\end{figure}

\Cref{fig:qualitative} shows a representative axial slice (slice~70 out of 150) from a test volume, juxtaposing the four MRI modalities, the ground-truth tumor masks, and the corresponding Active-DiNTS predictions.

The slice picked here is a mid-brain region where tumor density is high and lesion boundaries are reasonably well delineated across modalities. The Edema mask (channel~1) is reproduced with high fidelity, both in overall extent and in subtler peripheral patterns. The Non-Enhancing core (channel~0) is also delineated accurately. The Enhancing tumor (channel~2), being the smallest and most irregularly shaped subregion, is captured with reasonable precision, though fine boundary details are occasionally smoothed. This matches the quantitative trend: Enhancing is the hardest subregion across all methods, which is consistent with its limited spatial footprint and non-convex geometry. One slice is an illustration rather than evidence, and the quantitative tables remain the basis for any claim about accuracy.

\subsection{Ablation and Practical Notes}
\label{sec:ablation}

\paragraph{Uncertainty heuristic choice.}
Entropy, Variance, and Std converge to nearly identical final Dice values, with all pairwise differences under 0.5 points. We did not run significance tests across folds, so these differences should be treated as indistinguishable rather than as a ranking. That robustness simplifies hyperparameter selection in practice, since any of the three heuristics is a sensible default. In terms of memory overhead, Entropy was the cheapest of the three in our runs, which is consistent with it acting directly on class-level probabilities whereas Variance and Std need per-voxel deviations from the mean.

\paragraph{AL setup.}
The joint configuration delivers the fastest convergence overall, but topology-only is a close runner-up and can be preferable when the compute budget for weight updates is constrained. Weights-only stalls past the mid-process mark, reinforcing the message that the topology has to be actively refined alongside the weights.

\paragraph{Outlier sensitivity.}
Because Standard Deviation in particular responds to extreme values, volumes with severe imaging artifacts can score as highly uncertain without being informative about tumor structure. In a deployment setting this argues for screening such volumes before they consume oracle time. We did not run an ablation isolating the effect of such a filter on the learning curves, so we report this as a design consideration rather than as a measured result.

\paragraph{Positioning versus transformer and foundation-model approaches.}
Recent transformer-based 3D backbones~\cite{hatamizadeh2022unetr,hatamizadeh2022swin} and foundation-model adapters~\cite{ma2024segment,wu2025medical} operate at a different point in the resource trade-off curve. They rely on large-scale self-supervised pretraining or on prompting infrastructure at inference time. Active-DiNTS sits in the orthogonal regime in which neither prerequisite is available. With no pretraining corpus, no oracle prompts, and only a single GPU, it still has to discover a competitive architecture under a constrained annotation budget. Within that regime, and among the published results we are aware of on MSD Task01 BrainTumour, these are the strongest figures we have found. The two lines of work should be read as complementary; combining Active-DiNTS's bi-level AL loop with a transformer backbone or with foundation-model initialization is a promising direction that we revisit in Sec.~\ref{sec:future}.

\paragraph{FLOP budget.}
The FLOP gap between Active-DiNTS and NAS baselines deserves a final clarifying note. The reported values reflect training and search-time FLOPs of the dense, fully connected topologies favored by the AL-driven search at $\sigma{=}0.8$; they do not represent a hard constraint on deployment, since standard post-search compression (structured pruning, low-rank factorization, quantization) can in principle recover a leaner inference graph from the same set of weights. In our target setting, research and clinical workstations with constrained annotation budgets, the dominant practical cost is annotation, not FLOPs, and it is precisely on that axis that Active-DiNTS delivers its largest gains.

\section{Conclusions and Future Work}
\label{sec:future}

We have presented Active-DiNTS, a framework that brings together differentiable network topology search and pool-based Active Learning for 3D medical image segmentation. By using two interconnected and nested optimization loops, one for network weights and one for the macro/micro topology, and feeding both from uncertainty-ranked labeled pools, Active-DiNTS finds competitive U-Net-style architectures while requiring far fewer annotations and a much smaller GPU footprint than mainstream NAS pipelines.

On MSD Task01 BrainTumour, Active-DiNTS exceeds the average Dice of DiNTS, C2FNAS, nnU-Net, and other strong baselines, with the most pronounced gains on Edema (about 10\,pp) and Non-Enhancing core (about 5\,pp). It does so on a single GPU and can match the eight-GPU DiNTS baseline in as little as 0.21 GPU-days. The architectures it produces are denser and more FLOP-heavy, yet their parameter counts and memory footprints remain in line with prior work, and the removal of a separate retraining stage further compresses the pipeline.

Four caveats accompany our results. First, voxel-wise uncertainty heuristics are sensitive to class imbalance and can be misled by outliers, which is a known weakness of this family of acquisition functions and one we did not quantify here.
Second, the dense topologies favored by the search translate into high training-time FLOPs that, while acceptable during search, may need post-hoc compression for latency-sensitive deployment.
Third, our evaluation concentrates on a single 3D MRI benchmark; this choice was deliberate, in order to enable direct and controlled comparisons against the DiNTS reference and prior NAS works that all report on the same task, but extending the analysis to additional MSD tasks and to non-MRI modalities is a natural next step.
Fourth, baseline numbers are taken from the original publications rather than reproduced under our own pipeline, so the comparison carries whatever protocol differences those works embed.

As future work, we will extend Active-DiNTS along four directions: (i)~\emph{wider benchmarking}, including additional MSD tasks (Pancreas, Heart, Prostate) and 2D natural-image segmentation (COCO, PASCAL VOC) to probe cross-task and cross-domain transfer; (ii)~\emph{richer AL strategies}, such as stream-based selection, query synthesis, or committee-based disagreement, which may yield more sample-efficient querying; (iii)~\emph{deployment-aware compression}, via structured pruning and quantization, to bring the inference cost of dense Active-DiNTS networks into line with resource-constrained clinical hardware; and (iv)~\emph{integration with foundation-model backbones}~\cite{kirillov2023segment,ravi2024sam2,ma2024segment,hatamizadeh2022unetr,shu2026efficient}, layering the bi-level AL loop on top of pretrained transformer or SAM-style models to combine annotation efficiency with the representational power of large-scale pretraining.

\section*{Data and Code Availability}
The data used in this study are publicly available (MSD/Task01 BrainTumour). Source code is available at \url{https://github.com/geantrindade/Active-DiNTS}.

\section*{Relation to Prior Work}
This paper develops and extends material from the first author's doctoral thesis, \emph{Meta-Learning applied to Neural Architecture Search: Towards new interactive learning approaches for indexing and analyzing images from expert domains} (University of S\~ao Paulo / La Rochelle Universit\'e, 2024). The experimental results reported here were obtained in that work.

\section*{Declaration of Competing Interest}
The authors declare no known competing financial interests or personal relationships that could have appeared to influence the work.

\section*{Acknowledgments}
This study was financed by Brazilian and French agencies: FAPESP (2019/19994-2, 2020/09835-1), CNPq (422705/2021-7, 309858/2021-6), CAPES (614/2020), and Nouvelle-Aquitaine Region project ANIMONS (2017-1R50112-17004376-00013425-THESE).

\section*{Declaration of Generative AI and AI-Assisted Technologies in the Manuscript Preparation Process}
During the preparation of this work, the authors used Claude (Anthropic) and Gemini (Google) to assist with English-language editing, literature retrieval, and manuscript preparation. After using these tools, the authors reviewed and edited the content as needed and take full responsibility for the content of the published article.

\bibliographystyle{unsrt}
\bibliography{references}

\end{document}